\PassOptionsToPackage{table,dvipsnames}{xcolor}
\documentclass[sigconf]{acmart}
\AtBeginDocument{%
  }

\copyrightyear{2026}
\acmYear{2026}
\setcopyright{cc}
\setcctype{by}
\acmConference[AI Summit '26]{ACM AI Summit 2026}{August 31-September 02, 2026}{Atlanta, GA, USA}
\acmBooktitle{ACM AI Summit 2026 (AI Summit '26), August 31-September 02, 2026, Atlanta, GA, USA}
\acmDOI{10.1145/3806096.3844882}
\acmISBN{979-8-4007-2638-5/2026/08}

\usepackage{array,tabularx,multirow,makecell,bm,mathtools,nicefrac,enumitem}
\usepackage{tcolorbox}
\tcbuselibrary{listings,breakable,skins}
\definecolor{ImproveOrange}{RGB}{255,230,200}
\definecolor{NoChangeGray}{RGB}{235,235,235}
\definecolor{DecreaseBlue}{RGB}{220,235,250}

\begin{document}

%%
%% The "title" command has an optional parameter,
%% allowing the author to define a "short title" to be used in page headers.

\title{Towards Explainable Benchmarking for Data-driven Post-Wildfire Debris Flow Prediction}

\author{Zhisheng Qi}
\email{Zhisheng.Qi@uga.edu}
\affiliation{%
  \institution{University of Georgia}
  \city{Athens}
  \state{Georgia}
  \country{USA}
}

\author{Li Zhu}
\email{ryanlizhu@uga.edu}
\affiliation{%
  \institution{University of Georgia}
  \city{Athens}
  \state{Georgia}
  \country{USA}
}

\author{Utkarsh Sahu}
\email{utkarsh@uoregon.edu}
\affiliation{%
  \institution{University of Oregon}
  \city{Eugene}
  \state{Oregon}
  \country{USA}
}

\author{Douglas Tommey}
\email{drt@uoregon.edu}
\affiliation{%
 \institution{University of Oregon}
 \city{Eugene}
  \state{Oregon}
 \country{USA}}

\author{Josh Roering}
\email{jroering@uoregon.edu}
\affiliation{%
 \institution{University of Oregon}
 \city{Eugene}
  \state{Oregon}
 \country{USA}}

\author{Yu Wang}
\email{yu.wang6@uga.edu}
\affiliation{%
  \institution{University of Georgia}
  \city{Athens}
  \state{Georgia}
  \country{USA}}
  
%%
%% By default, the full list of authors will be used in the page
%% headers. Often, this list is too long, and will overlap
%% other information printed in the page headers. This command allows
%% the author to define a more concise list
%% of authors' names for this purpose.
\renewcommand{\shortauthors}{Qi et al.}

\begin{abstract}
Post-wildfire debris flows (PFDFs) are destructive sediment-laden hazards triggered when intense rainfall strikes recently burned terrain, destabilizing hillslopes and threatening infrastructure, local economies, and community safety.
Data-driven methods have been proposed to learn predictive patterns directly from historical PFDF observations. However, the current research landscape of data-driven PFDF prediction remains highly fragmented across feature spaces, model architectures, and evaluation protocols, making rigorous comparison and the derivation of scientific insights difficult. 
Moreover, existing studies lack a systematic investigation into the relative importance of heterogeneous factors (e.g., meteorological conditions, terrain characteristics, soil properties, and burn severity) in triggering PFDF.
To address these limitations, we present unified benchmark for data-driven PFDF prediction, enabling fair and comprehensive evaluation across diverse models and feature configurations. Furthermore, to better understand underlying drivers of PFDF formation, we propose a reinforcement learning–based feature selection framework that identifies factors whose perturbations render positive and negative events indistinguishable, discovering the regional underlying mechanisms of PFDF occurrence across regions. Our code and benchmark are publicly available at \url{https://github.com/KINDLab-Fly/PFDF-Benchmark}.

% the underlying working mechanisms and 

% To further alleviate the black-box nature of purely learning-based predictions, we provide interpretable insights into the dominant drivers and governing mechanisms underlying PFDF formation.
%
% \textcolor{orange}{Extensive experiments demonstrate the effectiveness of benchmark and the feature importance.}
\end{abstract}



%%
%% Keywords. The author(s) should pick words that accurately describe
%% the work being presented. Separate the keywords with commas.

%%No need for manuscript revision information
% \received{20 February 2007}
% \received[revised]{12 March 2009}
% \received[accepted]{5 June 2009}

%% This command processes the author and affiliation and title
%% information and builds the first part of the formatted document.

\maketitle

\section{Introduction}
Postfire debris flows are highly destructive secondary hazards triggered by intense rainfall over recently burned terrain~\cite{staley2017prediction, sousa2026influence, sanders2025influence}, where fire-induced changes such as reduced infiltration capacity, hillslope destabilization, and increased sediment availability can rapidly generate high-energy debris flows following heavy precipitation~\cite{kern2017ml_pfdf, nikolopoulos2018pfdf_rf, movasat2026mudflow_ml}. These events pose severe threats to human life, critical infrastructure, and local economies~\cite{selander2024inventory}.
Therefore, accurately predicting PFDF occurrence and understanding its key driving factors are essential for understanding PFDF formation mechanisms, effective risk assessment, and hazard mitigation.

\begin{figure}[t!]
     \centering
     \vspace{4ex}
     \includegraphics[width=1\linewidth]{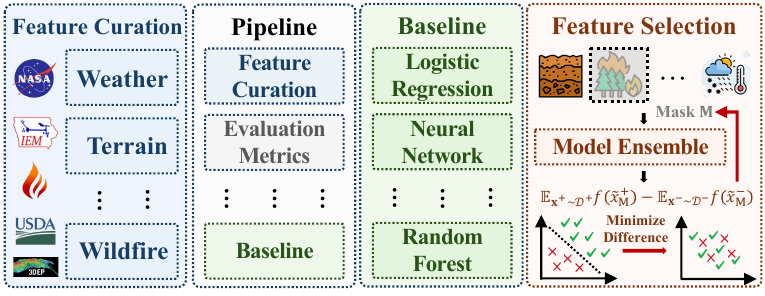}
     \vspace{-5ex}
     \caption{Benchmark Framework of Data-driven PFDF Prediction, including data curation, evaluation metrics, baseline, and feature selection. Heterogeneous environmental inputs support standardized model comparisons, while reinforcement learning identifies influential features.}
    \vspace{-6ex}
     \Description{Benchmark Framework of Data-driven PFDF Prediction, including data curation, evaluation metrics, baseline, and feature selection. Heterogeneous environmental inputs support standardized model comparisons, while reinforcement learning identifies influential features.}
     \label{fig:framework}
\end{figure}

% \begin{table}[t!]
% \centering
% \scriptsize
% \setlength{\tabcolsep}{3pt}
% \resizebox{\linewidth}{!}{%
% \begin{tabular}{lcccccc}
% \toprule
% \multirow{2}{*}{\textbf{Method}} & \multicolumn{4}{c}{\textbf{Feature factor}} & \multirow{2}{*}{\textbf{Dataset}} & \multirow{2}{*}{\textbf{Metric}}\\
% \cmidrule(lr){2-5}
% & \textbf{Wildfire} & \textbf{Soil} & \textbf{Terrian} & \textbf{Weather} & \\
% \midrule
% \cite{kern2017ml_pfdf} & \cmark & \cmark & \cmark & \cmark &  \\
% \cite{nikolopoulos2018pfdf_rf} & \cmark & \cmark & \cmark & \cmark &  \\
% \cite{orland2020hydrologic_dl} & \xmark & \cmark & \xmark & \cmark &  \\
% \cite{tang2023discharge_thresholds} & \cmark & \xmark & \cmark & \cmark &  \\
% \cite{movasat2026mudflow_ml} & \cmark & \cmark & \cmark & \cmark &  \\
% \textbf{Ours} & \cmark & \cmark & \cmark & \cmark & \\
% \bottomrule
% \end{tabular}
% }
% \caption{Prior data-driven or physics-aware studies related to post-wildfire debris-flow prediction and the major physical factors represented in their inputs. Check marks indicate that the study explicitly models or measures the factor.}
% \label{tab:prior_work}
% \vspace{-6ex}
% \end{table}

Existing PFDF prediction work moves from empirical rainfall thresholds and fixed logistic-regression formulations toward broader data-driven models. Classical machine-learning studies showed that nonlinear classifiers can extract useful signal from rainfall, burn severity, soil, and basin morphology: \cite{kern2017ml_pfdf} compared multiple classifiers on intermountain western-US burn-basin data, while \cite{nikolopoulos2018pfdf_rf} found that random forests improved over rainfall thresholds and logistic regression on the western-US USGS database. More recent work has begun to reconnect learning with process understanding. \cite{tang2023discharge_thresholds} combined process-based runoff simulation with support vector machines to define post-fire discharge thresholds, and \cite{movasat2026mudflow_ml} used controlled flume experiments to study how rainfall intensity, slope, grain size, and soil hydrophobicity affect post-wildfire mudflow onset. Related landslide-warning research shows that deep learning can forecast hydrologic response from rainfall and soil moisture.

% , and pore-pressure observations~\cite{orland2020hydrologic_dl}. 
% Together, these studies establish the value of data-driven PFDF analysis, but they also leave a gap: prior work rarely combines field-scale.
% FDF benchmarking, operational.
% forecast-text context, curated soil--terrain--fire augmentation.
% , and physically constrained neural interactions in one unified framework.

Despite recent progress in data-driven PFDF prediction, existing studies remain fragmented across feature settings, model architectures, and evaluation protocols, making rigorous and fair comparison across methods difficult. In addition, most existing approaches rely on rainfall-specific statistics while overlooking the richer meteorological context and storm dynamics in operational weather forecast reports\footnote{\protect\url{https://www.weather.gov/lox/\#}}. Furthermore, the black-box nature of data-driven prediction, especially deep learning methods, limits interpretable scientific understanding of the mechanisms underlying PFDF formation.
To address these limitations, we present the first explainable benchmark for data-driven PFDF prediction. Our framework systematically incorporates physical and meteorological features as augmented contextual information and further introduces a reinforcement learning–based feature selection framework to identify globally and regionally important driving factors associated with PFDF occurrence. Our contributions are as follows:
%
% \newpage
\begin{itemize}[leftmargin=*, itemsep=0pt]
\item \textbf{Unified Data-driven PFDF Prediction Benchmark.} We review existing data-driven PFDF prediction methods and establish a unified benchmark for fair comparison and scientific insight derivation across diverse approaches.

\item \textbf{Weather Feature Augmentation via LLM Processing.} We leverage LLM-based text processing to curate operational weather forecasts and USGS elevation products, extracting storm and terrain contextual signals to augment PFDF prediction.

\item \textbf{Important Feature Selection.} We propose a feature selection framework based on reinforcement learning that identifies influential drivers by minimizing predictive distinctions between positive and negative groups, revealing key global and region-specific factors underlying PFDF formation.
\end{itemize}

\section{Benchmark Setup}\label{sec-benchmark}
We study binary post-wildfire debris-flow (PFDF) occurrence prediction at the storm-watershed-event level~\cite{staley2016}. Each sample corresponds to a paired observation of a burned watershed segment and a rainfall storm event ${(\mathbf{x}_i, y_i)}$. For each post-fire storm affecting a monitored burned watershed segment, the dataset records storm forcing conditions, watershed and burn attributes, and a binary label $y_i \in \{0,1\}$ indicating whether a debris flow was observed during the storm response. The input feature $\mathbf{x}_i$ consists of meteorological features
(e.g., rainfall intensity and storm duration), wildfire features (e.g., burn severity and dNBR), and geologic/geomorphic
features (e.g., terrain slope and soil erodibility).

\subsection{Design Space} 

\subsubsection{Dataset.} We leverage the post-fire debris-flow database released by~\cite{staley2016}, where each entry represents a monitored storm response for a burned watershed segment, including both debris-flow and non-debris-flow outcomes. The dataset contains 1,243 storm-watershed events from 33 fires across 10 geographic zones in the western United States, with 316 positive cases (25.4\%).
For consistent evaluation, we use a complete-case subset by removing events with missing sub-hourly rainfall variables required by baseline models, yielding 1,131 events with 292 positives (25.8\%). Each event includes storm summaries, burn severity, soil and terrain attributes, and spatial-temporal indicators.

\subsubsection{Feature Setup.}
The original feature table follows the USGS post-fire debris-flow database~\cite{staley2016}. Each storm-watershed observation contains a binary debris-flow response with a total of 26 attributes spanning meteorology, wildfire, and geology~\cite{staley2016updated, gartner2014empirical}. 
% Storm variables characterize rainfall forcing through duration, accumulation, average intensity, and short-window peak rainfall statistics at 15, 30, and 60 minutes. Burn and watershed attributes include contributing area, dNBR-based burn severity, steep high-severity burn proportion, soil erodibility (KF factor), geographic zone, storm month, and rain-gauge distance. 
These variables form the core feature for the USGS-deployed model.

We further augment the baseline event attributes by linking each record to additional physical and meteorological context. Terrain attributes (e.g., mean slope and aspect), are derived from USGS 3DEP elevation products based on the geo-coordinates of each event\footnote{\url{https://www.usgs.gov/3d-elevation-program}}. Antecedent hydrologic conditions are represented by NLDAS-2 soil-moisture summaries and time since fire. In addition, we match storm events to National Weather Service forecast products\footnote{\url{https://mesonet.agron.iastate.edu/api/}}, including Area Forecast Discussions and Hazardous Weather Outlooks. We then prompt an LLM (Table~\ref{tab:prompt_weather_example}) to convert these narrative forecasts into structured PFDF weather indicators, such as precipitation threat, extreme rainfall potential, and overall PFDF risk.
% Overall, the augmented table extends the original USGS features with terrain, hydrologic recovery, and forecast-derived meteorological information.

\subsubsection{Baseline and Model Families.}\label{sec:baselines}
Following prior studies~\cite{staley2016, kern2017ml_pfdf, nikolopoulos2018pfdf_rf}, we benchmark a fixed operational reference model and a suite of trainable learning baselines. Physics logistic regression (P-LR) denotes the published USGS operational model with frozen coefficients~\cite{staley2016}. The trainable baselines include logistic regression (LR), classification and regression tree (CART)~\cite{kern2017ml_pfdf}, support vector machine (SVM)~\cite{kern2017ml_pfdf}, neural network (NN)~\cite{kern2017ml_pfdf}, random forest (RF)~\cite{nikolopoulos2018pfdf_rf}, and histogram gradient boosting (HGB). For each baseline, we also evaluate an augmented-feature variant with terrain attributes, antecedent hydrologic conditions, and LLM-processed PFDF weather indicators, denoted by $\mathrm{LR}^{*}$, $\mathrm{CART}^{*}$, $\mathrm{SVM}^{*}$, $\mathrm{NN}^{*}$, $\mathrm{RF}^{*}$, and $\mathrm{HGB}^{*}$.

\subsubsection{Evaluation Metric.} Our benchmark computes a comprehensive set of binary-classification metrics,
including accuracy, precision, recall, false-alarm behavior, and Threat Score~\cite{staley2016, kern2017ml_pfdf, nikolopoulos2018pfdf_rf}. Following PFDF evaluation used for USGS deployment model~\cite{staley2016}, we report Threat Score (TS): $\mathrm{TS}=\frac{\mathrm{TP}}{\mathrm{TP}+\mathrm{FP}+\mathrm{FN}}$. TS is well-suited to imbalanced hazard-detection problems because it directly
rewards correct debris-flow detections while penalizing both missed events (false negatives) and false alarms (false positives), without allowing the large number of true negatives to dominate the evaluation.

\begin{table}[t!]
\centering
\caption{The LLM is constrained to fill a predefined JSON schema with 18 numeric PFDF-relevant weather indicators and one concise textual summary. These fields decompose broad forecast signals, including rainfall intensity, storm duration or repetition, flash-flood language, convection, terrain enhancement, antecedent wetness, burn-scar sensitivity, and overall PFDF weather risk.}
\label{tab:prompt_weather_example}
\scriptsize
\begin{tcolorbox}[
    colback=white,
    colframe=green!40!black,
    boxrule=1pt,
    title={LLM-based Processing to Structure Area Forecast Discussion Text Corpus},
    colbacktitle=green!40!black,
    coltitle=white
]

\textbf{Input AFD:} \texttt{AFDLOX@2009-12-09T22:01}

\textbf{Forecast excerpt:}
``Relatively cold temperatures will continue tonight, with slight warming expected
through the week. A storm system will bring a chance of light to moderate rain on
Thursday into early Friday. Another storm system is expected to affect the area
over the weekend, with the potential for more significant rainfall. This storm
produces significant orographic rainfall, with preliminary totals of 0.5 to 1.5
inches for coastal and valley areas and 1.5 to 2.5 inches in the foothills and
mountains.''

\textbf{Extraction instruction:}
Read the AFD as an operational meteorologist and post-wildfire debris-flow hazard analyst. Convert only the PFDF-relevant weather signal into a compact and structured record. Use only the information stated in the forecast text.

\textbf{Structured output:}

\emph{Summary:} ``Two storm systems are expected with moderate to significant
rainfall, especially in foothills and mountains due to orographic enhancement.''

\begin{center}
\scriptsize
\setlength{\tabcolsep}{6pt}
\renewcommand{\arraystretch}{1.05}
\resizebox{\linewidth}{!}{%
\begin{tabular}{@{}>{\ttfamily}l>{\ttfamily}l@{}}
precipitation\_signal=1 & localized\_extreme\_signal=0 \\
short\_duration\_intensity\_signal=0 & mentions\_heavy\_rain=0 \\
storm\_duration\_signal=1 & mentions\_thunderstorm=0 \\
training\_repetition\_signal=0 & mentions\_flash\_flood=0 \\
antecedent\_wetness\_signal=0 & mentions\_burn\_scar\_or\_debris\_flow=0 \\
convective\_signal=0 & mentions\_training\_or\_repeating\_cells=0 \\
orographic\_enhancement\_signal=1 & mentions\_slow\_storm\_motion=0 \\
explicit\_flash\_flood\_signal=0 & mentions\_monsoon\_or\_tropical\_moisture=0 \\
burn\_scar\_sensitivity\_signal=0 & overall\_pfdf\_weather\_risk=1 \\
\end{tabular}%
}
\end{center}
\end{tcolorbox}
\end{table}

\subsubsection{Hyperparameter.}
We evaluate four zone stratified train/\allowbreak validation/\allowbreak test ratios:
50/25/25, 60/20/20, 70/15/15, and 80/10/10, using five split seeds. For each ratio and split seed, hyperparameters are grid searched by validation Threat Score over learning rate $\{10^{-4},\allowbreak 3{\times}10^{-4},\allowbreak 10^{-3},\allowbreak 3{\times}10^{-3}\}$,
dropout $\{0.0,0.1,0.2,0.3,0.5\}$, hidden width
$\{32,64,128,256,512\}$, and weight decay $\{0,10^{-4}\}$.
% For neural models, the validation Threat Score is averaged over initialization seeds 42-46 for hyperparameter selection, and the selected configuration is evaluated over initialization seeds 42-51. 
All neural models are trained with Adam, binary cross-entropy, and early stopping with patience 50 on validation Threat Score.

\subsection{RL-based Feature Selection}\label{sec-RL}
Given trained PFDF prediction models $\mathcal{F}$, we learn a binary feature-removal mask $\mathbf{M}\in\{0,1\}^{d}$ to identify features responsible for separating positive and negative PFDF events. Here, $\mathbf{M}_{j}=1$ suppresses feature $j$, while $\mathbf{M}_{j}=0$ retains it. Masked features are replaced by reference values: $\tilde{\mathbf{x}}_{\mathbf{M}}
= (\mathbf{1}-\mathbf{M})\odot \mathbf{x}
+ \mathbf{M}\odot \mathbf{b}$ where $\mathbf{b}$ is a baseline feature vector estimated from the training set. Standardized continuous features use $0$ (training-set mean after normalization), while the ordinal zone feature uses the most frequent training-zone value. For a model $f\in\mathcal{F}$, we measure the predictive separation between positive and negative samples under mask $\mathbf{M}$ using the standardized mean gap:
$D_f(\mathbf{M})=
\frac{
\left|
\mathbb{E}_{\mathbf{x}^{+}\sim \mathcal{D}^{+}}
f(\tilde{\mathbf{x}}^{+}_{\mathbf{M}})
-
\mathbb{E}_{\mathbf{x}^{-}\sim \mathcal{D}^{-}}
f(\tilde{\mathbf{x}}^{-}_{\mathbf{M}})
\right|
}{
\sqrt{
\mathrm{Var}_{\mathbf{x}^{+}\sim \mathcal{D}^{+}}
f(\tilde{\mathbf{x}}^{+}_{\mathbf{M}})
+
\mathrm{Var}_{\mathbf{x}^{-}\sim \mathcal{D}^{-}}
f(\tilde{\mathbf{x}}^{-}_{\mathbf{M}})
+\epsilon
}}$.
We then optimize masks that maximally collapse this separation relative to the unmasked input: $\mathbf{M}^{*}
= \arg\max_{\mathbf{M}}\mathbb{E}_{f\sim\mathcal{F}}
[1-\frac{D_f(\mathbf{M})}{D_f(\boldsymbol{0})}
] -\lambda \|\mathbf{M}\|_{0}$. This objective identifies features whose removal makes positive and negative predictions less distinguishable, implying their importance for PFDF discrimination, while the sparsity term avoids suppressing excessive features. We optimize the discrete masking objective using a REINFORCE-style policy gradient~\cite{liu2021automated, fan2020autofs, fan2021interactive, sutton1999policy}.

% \begin{table}[t!]
% \centering
% \caption{Dataset used for the controlled benchmark.}
% \label{tab-data}
% \begin{tabular}{lr}
% \toprule
% \textbf{Attribute} & \textbf{Value} \\
% \midrule
% Raw monitored storm--watershed events & 1,243 \\
% Positive responses, raw & 316 (25.4\%) \\
% Fires & 33 \\
% Geographic zones & 10 \\
% Complete-case benchmark events & 1,131 \\
% Positive responses, complete case & 292 (25.8\%) \\
% \bottomrule
% \end{tabular}
% \end{table}

% \begin{table}[t]
% \centering
% \caption{Benchmark model families and the comparison each one supports.}
% \label{tab:model_families}
% \small
% \begin{tabularx}{\linewidth}{p{1.8cm}X}
% \toprule
% \textbf{Model} & \textbf{Role in the benchmark} \\
% \midrule
% P-LR & Fixed USGS logistic-regression reference with frozen published coefficients. \\
% NN & Structured neural baseline using only tabular storm, watershed, burn, season, and zone inputs. \\
% NN+W & Tests whether dense pre-event forecast-text embeddings add predictive value. \\
% NN+SW & Tests whether compact structured weather indicators preserve forecast signal more efficiently. \\
% NN+AUG & Tests whether terrain, antecedent soil moisture, and recovery-time covariates improve the tabular baseline. \\
% NN+selected SW+AUG & Tests whether the strongest weather and physical covariates can be fused in a simple MLP. \\
% NN+G & Tests whether a physics-informed graph constraint improves over an unconstrained augmented MLP. \\
% \bottomrule
% \end{tabularx}
% \end{table}

\begin{table}[t!]
\centering
\caption{Baseline benchmark under original and selected augmented feature sets. Entries report the mean test set Threat Score across five zone-stratified split seeds.}
\label{tab:baseline_original_augmented_features}
\small
\setlength{\tabcolsep}{2pt}
\begin{tabularx}{\linewidth}{@{}Xcccc@{}}
\toprule
\textbf{Method} & \textbf{50/25/25} & \textbf{60/20/20} & \textbf{70/15/15} & \textbf{80/10/10} \\
\midrule
USGS~\cite{staley2016} 
& $0.388 \pm 0.031$ 
& $0.371 \pm 0.030$ 
& $0.376 \pm 0.030$ 
& $0.402 \pm 0.044$ \\
\midrule
LR~\cite{nikolopoulos2018pfdf_rf, kern2017ml_pfdf} 
& $0.449 \pm 0.022$ 
& $0.409 \pm 0.052$ 
& $0.441 \pm 0.035$ 
& $0.484 \pm 0.035$ \\
CART~\cite{kern2017ml_pfdf} 
& $0.523 \pm 0.032$ 
& $0.509 \pm 0.043$ 
& $0.556 \pm 0.078$ 
& $0.529 \pm 0.042$ \\
SVM~\cite{kern2017ml_pfdf} 
& $0.504 \pm 0.046$ 
& $0.464 \pm 0.034$ 
& $0.524 \pm 0.058$ 
& $0.552 \pm 0.066$ \\
RF~\cite{nikolopoulos2018pfdf_rf} 
& \underline{$0.585 \pm 0.027$} 
& $\bm{0.617 \pm 0.037}$ 
& $\bm{0.658 \pm 0.046}$ 
& $\bm{0.674 \pm 0.041}$ \\
HGB 
& $\bm{0.604 \pm 0.038}$ 
& \underline{$0.611 \pm 0.032$} 
& \underline{$0.640 \pm 0.040$} 
& \underline{$0.648 \pm 0.052$} \\
NN~\cite{kern2017ml_pfdf} 
& $0.540 \pm 0.030$ 
& $0.558 \pm 0.046$ 
& $0.583 \pm 0.032$ 
& $0.613 \pm 0.012$ \\
\midrule
LR$^{*}$ 
& $0.457 \pm 0.025$ 
& $0.483 \pm 0.022$ 
& $0.481 \pm 0.021$ 
& $0.483 \pm 0.033$ \\
CART$^{*}$ 
& $0.504 \pm 0.035$ 
& $0.542 \pm 0.044$ 
& $0.542 \pm 0.050$ 
& $0.549 \pm 0.030$ \\
SVM$^{*}$ 
& $0.547 \pm 0.020$ 
& \underline{$0.597 \pm 0.029$} 
& \underline{$0.648 \pm 0.025$} 
& \underline{$0.670 \pm 0.033$} \\
RF$^{*}$ 
& \underline{$0.600 \pm 0.037$} 
& $\bm{0.634 \pm 0.046}$ 
& $\bm{0.685 \pm 0.048}$ 
& $\bm{0.687 \pm 0.065}$ \\
HGB$^{*}$ 
& $0.595 \pm 0.062$ 
& $0.605 \pm 0.031$ 
& $0.634 \pm 0.031$ 
& $0.652 \pm 0.028$ \\
NN$^{*}$ 
& $\bm{0.604 \pm 0.021}$ 
& \underline{$0.611 \pm 0.041$} 
& $0.632 \pm 0.018$ 
& $0.663 \pm 0.043$ \\
\bottomrule
\end{tabularx}
\end{table}

\section{Experimental Results}

\subsection{Benchmarking Performance}

Table~\ref{tab:baseline_original_augmented_features} compares classical and neural baselines under both the original and the selected augmented feature table. The fixed USGS logistic-regression (USGS) model is included as an operational reference and is not retrained. All other methods follow the same repeated zone-stratified evaluation protocol, with entries reporting the mean test Threat Score and standard deviation across five split seeds. While the original feature table already provides strong performance for tree-based ensembles, the selected augmented features further improve most model families, particularly SVM, random forest, and neural network models. The widely deployed USGS logistic-regression baseline performs substantially worse than modern learning methods. This is primarily because its fixed coefficients cannot adaptively capture the diverse regional mechanisms underlying PFDF formation, an issue we analyze in Section~\ref{sec-analysis}.

\begin{figure}[t!]
     \centering
     \includegraphics[width=1\linewidth]{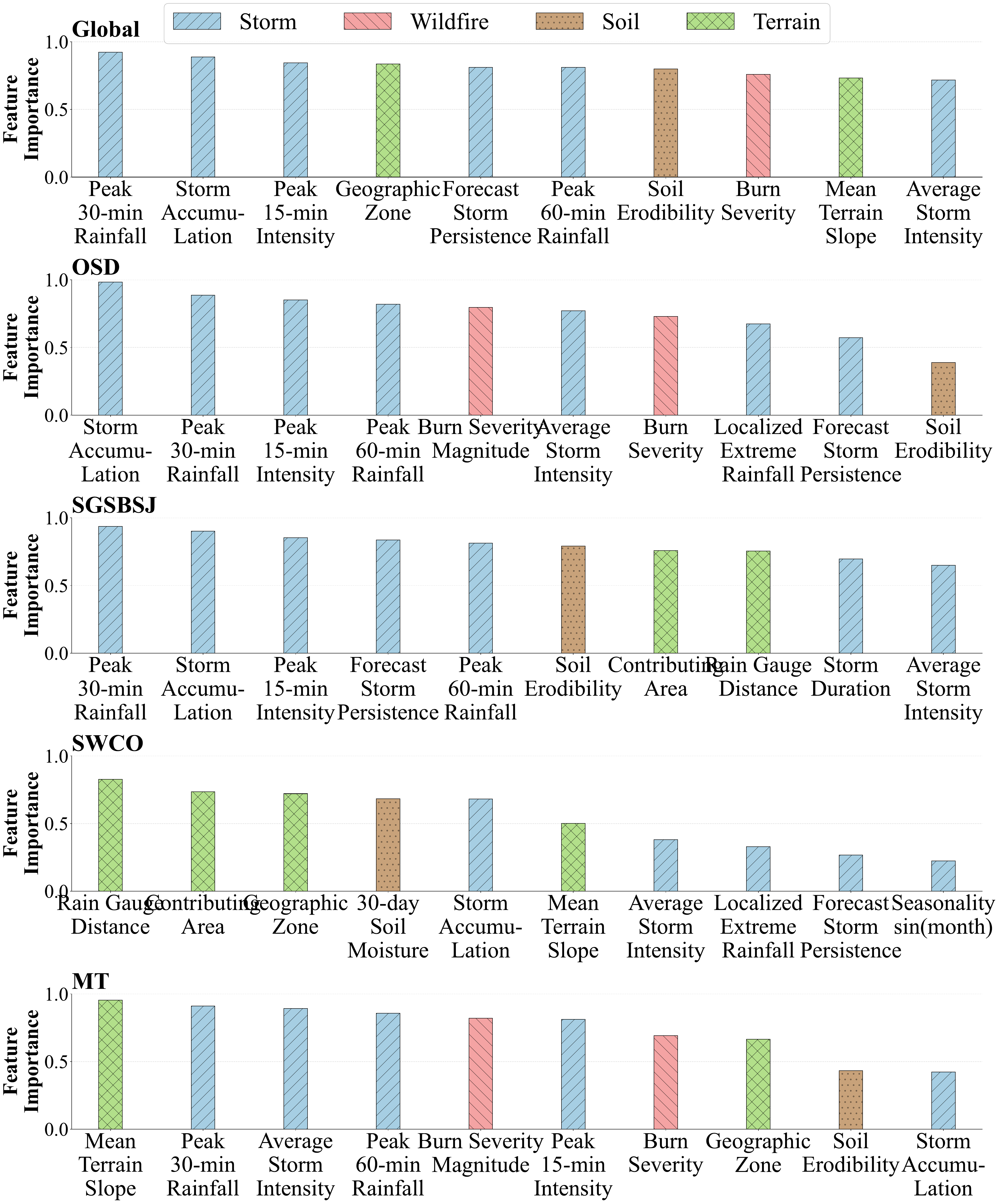}
     \caption{Top-10 Important Features for PFDF prediction across regions with color denoting feature category.}
\Description{Top-10 Important Features for PFDF prediction across regions with color denoting feature category.}
     \label{fig:feature_importance}
\end{figure}

\subsection{Global Feature Importance}
After applying the RL feature selection in Section~\ref{sec-RL}, we visualize the top-10 important features across different zones

From the global feature importance analysis, storm-related variables dominate the most influential features, aligning with prior literature identifying storm conditions as the primary triggering factor for PFDF events~\cite{staley2016}. Meanwhile, soil erodibility, burn severity, and terrain slope also consistently emerge as critical contributing factors. \textit{Notably, the geographical zone itself appears as a highly important feature, highlighting the strong region-specific characteristics underlying PFDF formation.} This observation suggests that different regions exhibit distinct interactions among meteorological, wildfire, soil, and terrain conditions, further motivating the need for adaptive and region-aware prediction models. This finding also resonates with substantially lower performance of the rigid fixed-coefficient USGS method under the zone-stratified evaluation setting in Table~\ref{tab:baseline_original_augmented_features}, as a single globally calibrated logistic formulation may struggle to capture region-dependent PFDF mechanisms.

Comparing different zones, we can clearly observe substantial variation in feature importance across regions. OSD and SGSBSJ, located in California, are primarily dominated by storm/rainfall-related factors, indicating the strong influence of meteorological triggering conditions in these regions. In contrast, for SWCO and WMT, terrain-related and geographical features emerge as the most influential factors, suggesting that topographical and regional characteristics play a more dominant role in determining PFDF occurrence in these areas.

\begin{table}[t!]
\centering
\caption{Region-wise absolute Threat Score improvement of the original-feature neural network over the fixed USGS logistic-regression reference. Entries report $\Delta\mathrm{TS}=\mathrm{TS}_{\mathrm{NN}}-\mathrm{TS}_{\mathrm{USGS}}$ averaged across split seeds. Region counts are reported as total records (positive/negative).}
\label{tab:region_nn_usgs_improvement}
\small
\setlength{\tabcolsep}{1.2pt}
\begin{tabular}{l|llcccc}
\toprule
\textbf{Area} & \textbf{Zone} & \textbf{Records} & \textbf{50/25/25} & \textbf{60/20/20} & \textbf{70/15/15} & \textbf{80/10/10} \\
\midrule
\multirow{3}{*}{\makecell[l]{\textbf{Southern} \\\textbf{California}}}
& OSD    & 103 (25/78)   & \cellcolor{ImproveOrange}$+0.19$ & \cellcolor{ImproveOrange}$+0.23$ & \cellcolor{ImproveOrange}$+0.22$ & \cellcolor{ImproveOrange}$+0.19$ \\
& SGSBSJ & 646 (176/470) & \cellcolor{ImproveOrange}$+0.16$ & \cellcolor{ImproveOrange}$+0.19$ & \cellcolor{ImproveOrange}$+0.19$ & \cellcolor{ImproveOrange}$+0.17$ \\
& VEN    & 24 (0/24)     & \cellcolor{NoChangeGray}$+0.00$ & \cellcolor{NoChangeGray}$+0.00$ & \cellcolor{NoChangeGray}N/A      & \cellcolor{NoChangeGray}$+0.00$ \\
\midrule
\multirow{2}{*}{\textbf{Arizona}}
& NAZ    & 71 (5/66)     & \cellcolor{NoChangeGray}$+0.05$ & \cellcolor{NoChangeGray}$+0.12$ & \cellcolor{NoChangeGray}$+0.00$ & \cellcolor{NoChangeGray}$+0.00$ \\
& SAZ    & 13 (2/11)     & \cellcolor{NoChangeGray}$+0.00$ & \cellcolor{NoChangeGray}$+0.00$ & \cellcolor{NoChangeGray}$+0.00$ & \cellcolor{NoChangeGray}$+0.00$ \\
\midrule
\textbf{N-Mexico}
& CNM    & 52 (35/17)    & \cellcolor{NoChangeGray}$-0.01$ & \cellcolor{NoChangeGray}$+0.00$ & \cellcolor{NoChangeGray}$+0.05$ & \cellcolor{NoChangeGray}$+0.06$ \\
\midrule
\multirow{3}{*}{\makecell[l]{\textbf{Colorado}\\\textbf{Western}\\\textbf{Plateau}}}
& FRCO   & 22 (4/18)     & \cellcolor{NoChangeGray}$-0.07$ & \cellcolor{NoChangeGray}$-0.20$ & \cellcolor{NoChangeGray}$+0.00$ & \cellcolor{NoChangeGray}$+0.00$ \\
& SWCO   & 15 (10/5)     & \cellcolor{ImproveOrange}$+0.30$ & \cellcolor{ImproveOrange}$+0.33$ & \cellcolor{ImproveOrange}$+0.50$ & \cellcolor{ImproveOrange}$+0.52$ \\
& WCO    & 145 (17/128)  & \cellcolor{ImproveOrange}$+0.16$ & \cellcolor{ImproveOrange}$+0.25$ & \cellcolor{ImproveOrange}$+0.15$ & \cellcolor{ImproveOrange}$+0.24$ \\
\midrule
\makecell[l]{\textbf{Northern}\\\textbf{Rockies}}
& MT     & 40 (18/22)    & \cellcolor{DecreaseBlue}$-0.19$ & \cellcolor{DecreaseBlue}$-0.17$ & \cellcolor{DecreaseBlue}$-0.14$ & \cellcolor{DecreaseBlue}$-0.18$ \\
\bottomrule
\end{tabular}
\end{table}
\subsection{Further Analysis}\label{sec-analysis}
\begin{figure}[t!]
\centering
\includegraphics[width=\linewidth]{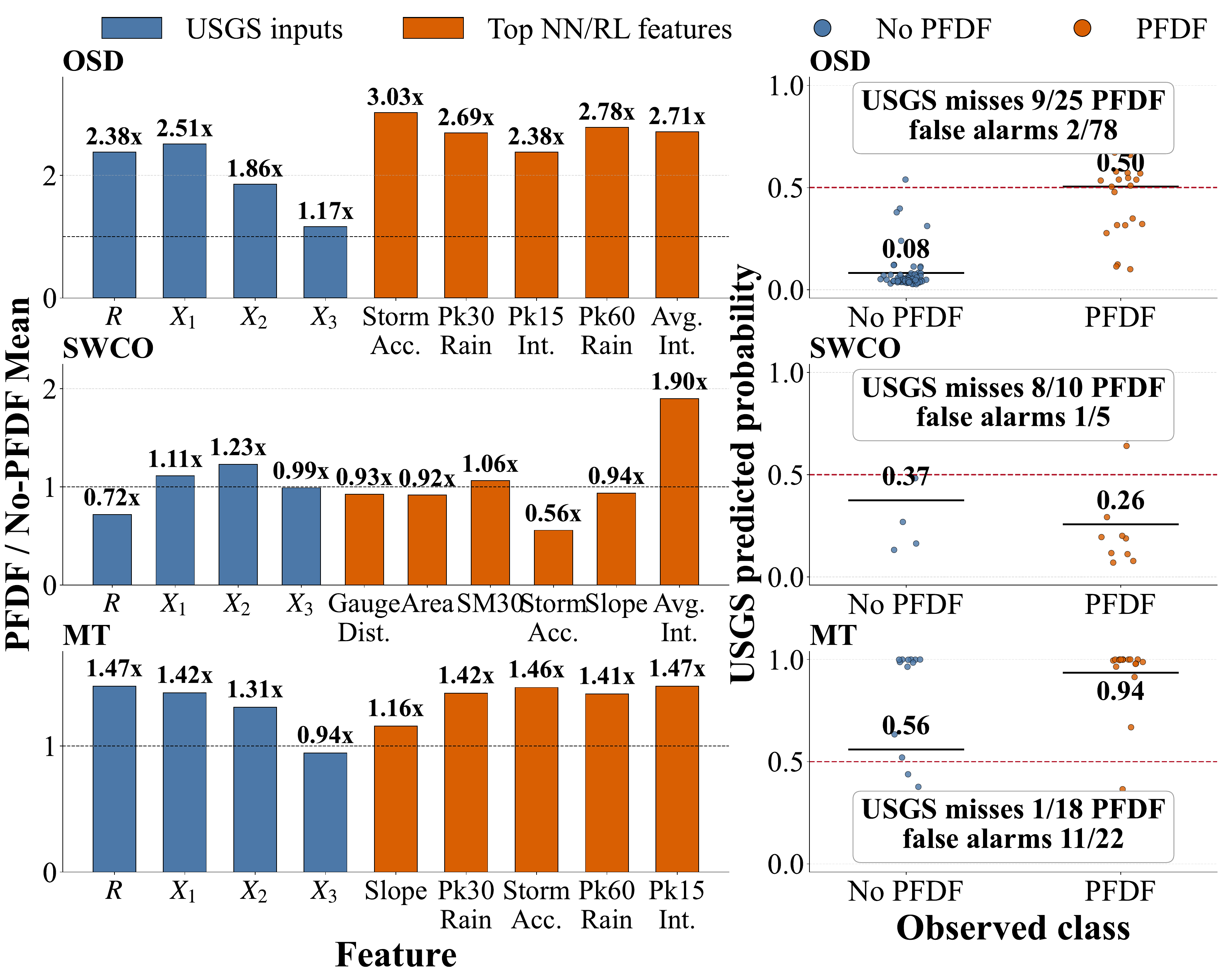}
\caption{Regional diagnostic for OSD, SWCO, and MT. Left: PFDF/non-PFDF mean ratios for USGS inputs and high-ranked learned features. Right: fixed USGS predicted probability by observed class, with the decision threshold and missed PFDF events.}
\Description{Regional diagnostic for OSD, SWCO, and MT. Left: PFDF/non-PFDF mean ratios for USGS inputs and high-ranked learned features. Right: fixed USGS predicted probability by observed class, with the decision threshold and missed PFDF events.}
\label{fig:regional_usgs_diagnostic}
\end{figure}

We further diagnose why the neural model improves over the fixed USGS logistic-regression reference in some regions but not others. Figure~\ref{fig:regional_usgs_diagnostic} compares three representative zones. In each row, the left panel reports the ratio between the PFDF-event mean and the non-PFDF-event mean for the USGS inputs ($R$, $X_1$, $X_2$, $X_3$) and high-ranked learned features from Figure~\ref{fig:feature_importance}; values above one indicate larger feature values in observed PFDF events. The right panel shows the fixed USGS probability distribution by observed class, together with the 0.5 decision threshold and the missed PFDF count. 
% This setting separates two questions: whether the fixed USGS variables already align with local PFDF occurrence, and whether additional learned features expose stronger regional separation.

\subsubsection{OSD/SGSBSJ} both belong to the broader Southern California and show similar improvement patterns. Figure~\ref{fig:regional_usgs_diagnostic} reports \texttt{OSD} as the representative case. In \texttt{OSD}, the neural model improves substantially over the fixed-coefficient USGS baseline because several non-USGS or expanded storm descriptors separate PFDF from non-PFDF events more strongly than the original formula terms. For example, storm accumulation, 30/60-minute rain, 15-minute intensity, and average storm intensity all show large PFDF/non-PFDF ratios, exceeding several of the original USGS inputs. The fixed formula, therefore, misses 9 of 25 PFDF events, leaving room for the neural model to learn additional storm-response structure.
% beyond the four hand-designed USGS terms.

\subsubsection{SWCO} is the clearest failure mode for the fixed USGS rule. The USGS equation assumes that larger rainfall forcing should monotonically increase PFDF probability~\cite{staley2016, staley2016updated, staley2017prediction}, but in this zone, the observed PFDF events have a smaller mean $R$ than non-PFDF events, violating this monotonic trend. This reversed local relationship causes the fixed model to rank many PFDF events too low, missing 8 of 10 positives. The neural model can partially correct this by using the zone representation and other covariates to learn a region-specific relationship, including cases where rainfall-related effects are weaker or negative relative to the global rule. This interpretation is consistent with Figure~\ref{fig:feature_importance}, where geographic zone is highly ranked for \texttt{SWCO}, suggesting that location helps the model adapt the mechanism rather than applying one universal coefficient pattern. This is consistent with distinctive PFDF triggering conditions deduced from physics-based models. 

\subsubsection{MT} illustrates the opposite regime. Here, the fixed USGS variables already align well with the observed outcome: $R$, $X_1$, and $X_2$ are larger for PFDF events than for non-PFDF events, while $X_3$ is close to parity. As a result, the fixed model misses only 1 of 18 PFDF events. Because the baseline already captures most positives in this region, there is less capacity for a flexible neural model to improve, and additional learned interactions can slightly reduce performance under the repeated split setting.

Overall, Figure~\ref{fig:regional_usgs_diagnostic} shows that regional PFDF mechanisms are not uniform. Some regions benefit from additional storm and susceptibility features, some require an adaptive correction to the sign or strength of the rainfall relationship, and some are already well described by the fixed USGS structure. This supports the need for models that empirically learn region-dependent mechanisms.

\section{Conclusion and Future Work}
This paper presents an explainable benchmark for data-driven post-wildfire debris-flow prediction. We found that learned models consistently outperform the fixed operational USGS logistic-regression baseline~\cite{staley2017prediction}, and augmented physical and meteorological features further improve predictive performance~\cite{li2025typhoformer}.
Beyond prediction, our reinforcement learning–based feature selection provides interpretable insights into PFDF drivers. Storm-related variables remain globally dominant, but feature-importance patterns vary substantially across regions, revealing distinct regional PFDF mechanisms. These findings indicate that globally fixed coefficients are insufficient across diverse geographic zones, whereas adaptive learning models can better capture region-specific dynamics while remaining physically interpretable.
Future work will move beyond influential-factor analysis toward mechanism discovery by developing physics-informed equations~\cite{dahal2025towards} for regional PFDF prediction. We will also investigate the trustworthiness of debris-flow prediction systems, including their vulnerability to knowledge poisoning and data extraction attacks~\cite{qi2026benchmarking, qi2026rigorizing}. 

% Future work will extend this benchmark in three directions. First, broader multi-agency and post-2020 PFDF inventories could improve coverage of underrepresented regions and support stronger out-of-region generalization tests. Second, prospective evaluation with real-time weather forecasts and uncertainty quantification is needed before operational warning use. Third, future models should more tightly integrate physical constraints, regional adaptation, and interpretable feature attribution so that learned predictors remain both accurate and scientifically useful for hazard assessment.

% \begin{acks}
% This work is supported 
% \end{acks}

%%
%% The next two lines define the bibliography style to be used, and
%% the bibliography file.
\bibliographystyle{ACM-Reference-Format}
\bibliography{main}

%% If your work has an appendix, this is the place to put it.
%\appendix

% \clearpage
% \section{Online Resources}

% Nam id fermentum dui. Suspendisse sagittis tortor a nulla mollis, in
% pulvinar ex pretium. Sed interdum orci quis metus euismod, et sagittis
% enim maximus. Vestibulum gravida massa ut felis suscipit
% congue. Quisque mattis elit a risus ultrices commodo venenatis eget
% dui. Etiam sagittis eleifend elementum.

% Nam interdum magna at lectus dignissim, ac dignissim lorem
% rhoncus. Maecenas eu arcu ac neque placerat aliquam. Nunc pulvinar
% massa et mattis lacinia.

\end{document}